\documentclass[conference]{IEEEtran}
\IEEEoverridecommandlockouts
\usepackage{cite}
\usepackage{amsmath,amssymb,amsfonts}
\usepackage{graphicx}
\usepackage{textcomp}
\usepackage{microtype}
\usepackage{subcaption}
\usepackage{multirow}
\usepackage{makecell}
\usepackage{booktabs} 
\usepackage{threeparttable}
\usepackage{amsmath,amsfonts}
\usepackage[table]{xcolor}
\usepackage{tcolorbox}
\usepackage{array}
\usepackage{listings}
\usepackage[percent]{overpic} 
\usepackage{algorithm}
\usepackage{algpseudocode}
\usepackage{hyperref}
\usepackage{cleveref}
\usepackage{afterpage}
\usepackage{placeins}
\usepackage{float}
\graphicspath{{figures/}}

\def\BibTeX{{\rm B\kern-.05em{\sc i\kern-.025em b}\kern-.08em
    T\kern-.1667em\lower.7ex\hbox{E}\kern-.125emX}}
\begin{document}

\title{Leveraging LLMs to Automate Energy-Aware Refactoring of Parallel Scientific Codes
}

\author{
\IEEEauthorblockN{Matthew T. Dearing, Yiheng Tao, and Zhiling Lan}
\IEEEauthorblockA{\textit{Department of Computer Science, College of Engineering} \\
\textit{University of Illinois Chicago} \\
Chicago, USA \\
\{mdear2, ytao28, zlan\}@uic.edu}
\and
\IEEEauthorblockN{Xingfu Wu and Valerie Taylor}
\IEEEauthorblockA{\textit{Mathematics and Computer Science Division} \\
\textit{Argonne National Laboratory} \\
Lemont, USA \\
\{xingfu.wu, vtaylor\}@anl.gov}
}

\maketitle

\begin{abstract}
Large language models (LLMs) are increasingly used for generating parallel scientific codes, with a primary focus on generating functionally correct code. Recent work has focused on generating performant code, with an emphasis on its execution time. However, energy efficiency is now recognized as a critical objective, given the significant power demands of large scale compute systems.  This paper addresses the research question of whether LLMs can generate energy-efficient parallel scientific codes when guided by empirical execution feedback. To answer this question, we propose LASSI-EE, an automated LLM-based refactoring framework that generates energy-efficient parallel codes through a multi-stage, iterative approach integrating runtime power profiling, energy-aware prompting, self-correcting feedback loops, and an LLM-as-a-Judge agent for screening generated code. We evaluate LASSI-EE using twenty-two representative scientific benchmarks and applications on NVIDIA A100 and AMD MI100 GPUs. The results indicate an average energy reduction of 36\% for MI100 and 34\% for A100, across trials that produced passing energy-reducing refactorings.

\end{abstract}


\section{Introduction}





Applications developed for high-performance computing (HPC) drive discovery across scientific domains, but their parallel implementations are complex, performance-sensitive, and increasingly constrained by energy limits. Current exascale systems require between 24 and 39 MW of power~\cite{Top500}. As energy demands strain both computing infrastructure and the environment, pursuing energy efficiency opportunities across the HPC stack, from hardware and system software to scientific applications, has become essential. Further, it is recognized that HPC systems are complex, including heterogeneous components (e.g., CPUs, GPUs), complex memory stacks, and unique features such as mixed-precision computation. Generating high-quality parallel scientific codes that are energy efficient is a complex challenge.

Large language models (LLMs) are now widely used for code generation, and recent work has extended this capability to parallel scientific codes. Studies have shown that LLMs can generate functionally correct parallel codes~\cite{Nichols2024} and translate between parallel programming frameworks~\cite{LASSI, tehranijamsaz_coderosetta_2024, ranasinghe_llm-assisted_2025, zhu_semi-supervised_2024}. 
However, most LLM-based code generation efforts prioritize functional correctness, with limited attention to performance. Among the few works that consider performance, optimization is typically measured in terms of execution time rather than energy consumption. This leaves a critical gap in the use of LLMs for HPC, where energy efficiency is an increasingly dominant constraint.

In this work, we investigate \textbf{whether LLMs can discover energy-efficient implementations of parallel scientific codes when given structured guidance and empirical execution feedback, and whether their strategies adapt to application and architecture or default to generic patterns.} 
Addressing this question requires overcoming three key challenges.

\textit{Guiding LLMs with empirical execution feedback.} 
LLMs generate code based on patterns learned during training, without knowledge of how that code will execute on a specific target platform. 
Yet, energy consumption depends on complex interactions between runtime behavior and hardware that static analysis cannot reliably infer. 
Effective refactoring strategies therefore vary across applications and architectures, making one-shot prompting insufficient. Achieving energy-efficient code generation requires iterative refinement guided by feedback derived from actual execution.

\textit{Code screening at scale where formal verification is not tractable.} Formal verification methods~\cite{siegel-mironova-avrunin-clarke:2006:symbolic, DBLP:conf/fm/BlomH14, abadi_verifying_2019, siegel_civl_2015} can rigorously prove equivalence for transformations within specified limits but are computationally expensive and not readily applicable to the diverse refactorings produced by LLM-driven code generation. 
At the same time, manual inspection is not scalable for large numbers of generated candidates. A practical, automated screening mechanism is therefore needed to filter out incorrect or semantically divergent implementations.

\textit{Evaluating energy-aware generation under non-determinism.} 
LLM inference is non-deterministic, so the same prompt can produce different code on different runs. Evaluating energy savings under this scenario requires considering the distribution of outputs across multiple generation attempts, not single-run results. Existing metrics such as pass@$k$~\cite{chenevaluating2021} quantify correctness likelihood but do not capture expected energy savings. As a result, there is no principled way to evaluate energy-aware code generation methods under a given generation budget.


To address the three challenges, we propose LASSI-EE (\textbf{L}LM-based \textbf{A}utomated \textbf{S}elf-correcting pipeline for generating parallel \textbf{S}c\textbf{I}entific codes for \textbf{E}nergy \textbf{E}fficiency), an automated refactoring pipeline that produces energy-efficient parallel codes tailored to a target execution platform. 
LASSI-EE builds upon the LASSI framework~\cite{LASSI}, which showed that domain context, self-prompting, and iterative self-correction augment pre-trained LLMs for translating parallel code between CUDA and OpenMP. While LASSI focused on cross-language translation and relied on heuristic human review for code validation, LASSI-EE extends this approach to energy optimization with automated code evaluation. 

LASSI-EE implements the following empirical feedback mechanisms. The pipeline measures code execution runtime and instantaneous power directly on the target GPU during every iteration. Total energy is calculated from these measurements and incorporated into successive prompts based upon whether the energy improved over the previous iteration. A context-building stage further provides platform documentation, identifies inefficiencies in the source code, and produces structured refactoring plans. For screening at scale, an LLM-as-a-Judge agent compares the source and refactored code along with their standard outputs from execution on identical inputs, flagging syntactic or semantic differences that suggest non-equivalent behavior. This screening filters out code candidates with execution behavior or an implementation that diverges from the source. The Judge supports large-scale screening but does not substitute for formal verification. To handle the non-determinism of LLM inference, we introduce \textit{energy-reduction@k}, a new metric that quantifies the expected energy reduction when generating $k$ code candidates and selecting the most efficient one.

We evaluate LASSI-EE on 22 representative applications from science and engineering domains across NVIDIA A100 and AMD MI100 GPUs, conducting 1,320 trials in total (22 applications $\times$ 30 runs per application $\times$ two platforms). Our results show that LLMs, when integrated into the LASSI-EE pipeline, can systematically generate energy-efficient parallel scientific code. Compared to the source code, LASSI-EE achieves average energy reductions of 36\% on AMD MI100 and 34\% on NVIDIA A100 across the trials in which it produces a passing energy-reducing refactoring, with consistent results across both GPU architectures. 

Further analysis reveals that the optimization strategies discovered by LASSI-EE are not generic but adapt to both application characteristics and hardware architecture.
Some applications benefit primarily from runtime reduction, while others achieve savings through power reduction or a combination. Across GPU platforms, only 3.5\% of optimization implementations are syntactically identical between AMD and NVIDIA generations, indicating that LASSI-EE produces device-appropriate optimizations rather than generic refactorings.
Overall, the key contributions of this research include:

\begin{itemize}


\item We develop \emph{LASSI-EE}, an LLM-based automated pipeline for 
energy-efficient parallel code generation that closes the loop between 
generation and measurement through iterative empirical feedback. LASSI-EE 
integrates runtime power profiling, energy-aware prompting, iterative 
self-correction, and an LLM-as-a-Judge screening agent that filters 
semantically inconsistent candidates by comparing source and refactored 
outputs on identical inputs, which complements, but does not replace, formal verification.

\item We demonstrate that LASSI-EE consistently generates energy-efficient parallel code across AMD MI100 and NVIDIA A100 GPUs, achieving an average of 36\% and 34\% energy reduction, respectively. We further demonstrate that its optimizations adapt to application and hardware characteristics, with energy savings decomposed into runtime and power contributions across memory hierarchy, algorithmic, device-specific, and parallelism strategies.

\item We introduce \emph{energy-reduction@$k$}, a new metric that quantifies expected energy reduction when generating $k$ code candidates and selecting the most efficient. We use it to characterize practical operating points for multi-attempt LASSI-EE deployment under the non-determinism inherent to LLM inference.

    
\end{itemize}


\section{LASSI-EE Framework}
\label{sec:method}

\begin{figure*}[!t]
\centering
\includegraphics[width=0.8\textwidth]{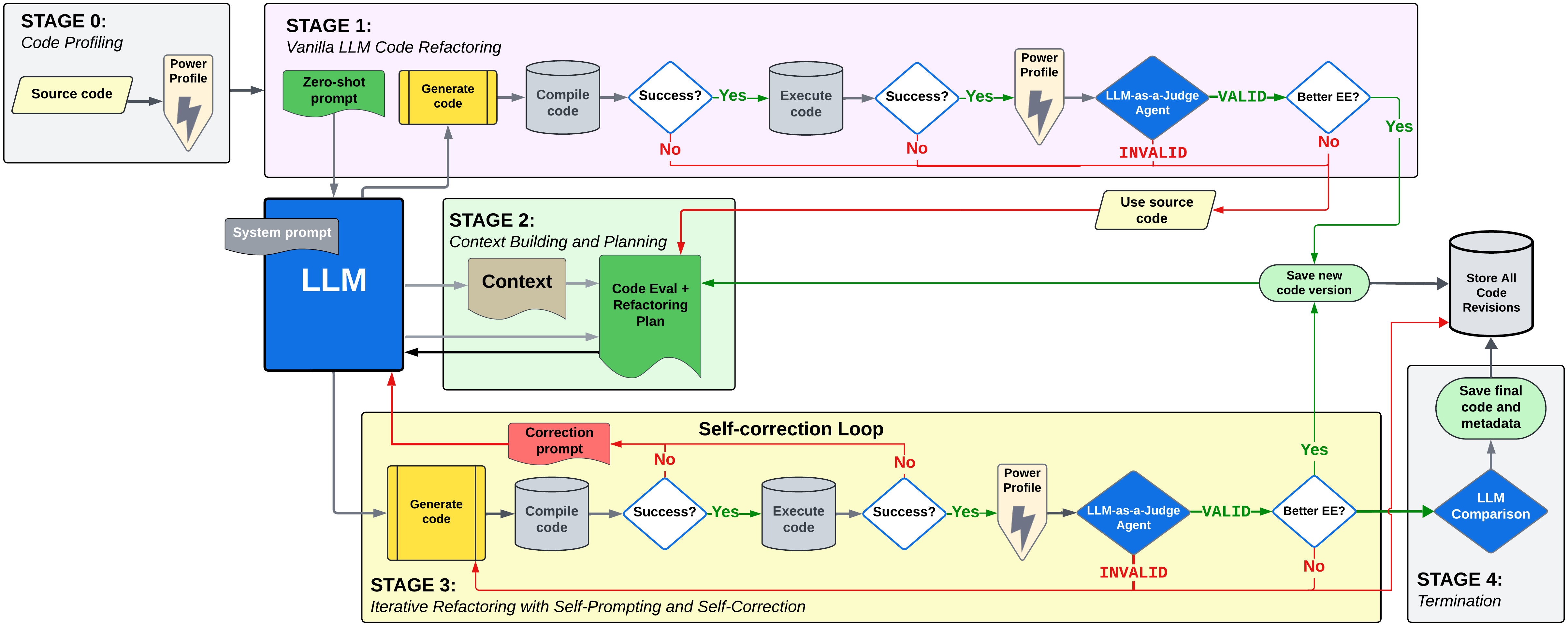}
\caption{
    Diagram of the \textbf{LASSI-EE pipeline}.}
\label{fig:lassiee-pipeline}
\end{figure*}

Figure~\ref{fig:lassiee-pipeline} illustrates the overall pipeline. Given a parallel code that runs on an accelerator device with a power-measurement capability, LASSI-EE proceeds through five stages. Stage 0 captures baseline performance and power profiling on the target system. Stage 1 produces a vanilla baseline refactoring with an out-of-the-box LLM. Stage 2 performs context-driven planning. Stage 3 conducts iterative self-prompted refactoring with self-correction. Stage 4 selects the best refactored version and terminates. The framework is \emph{LLM-agnostic} and \emph{extensible} to multiple architectures and programming models. All prompt templates used throughout the pipeline will be released in our open-source repository.


\subsection{Stage 0: Code Profiling}

The initial stage captures the baseline performance metrics of the parallel source code on the target system. The code is compiled, executed, and profiled for power consumption to establish a reference for comparison with refactored versions. Real-time power measurements are collected through platform-specific monitoring libraries (NVML on NVIDIA GPUs and ROCm-SMI on AMD GPUs) at 10-millisecond intervals during code execution and aggregated to calculate average power and total energy consumed. Unlike static analysis approaches relying on predicted costs, this provides empirical feedback for iterative refinement. Figure~\ref{fig:jacobi-power-profiles} shows the resulting power profile for the \texttt{jacobi} application alongside the profile of a variant refactored by LASSI-EE.




\subsection{Stage 1: Baseline Energy-Aware Refactoring}\label{sec:lassiee_1}

Stage 1 establishes a baseline by evaluating an out-of-the-box LLM’s inherent ability to generate energy-efficient code, serving as a reference for the gains from context-building, iterative refinement, and self-correction in later stages. We refer to this unassisted baseline as the \emph{vanilla LLM}.

A zero-shot prompt instructs the vanilla LLM to refactor the source code for energy efficiency without context or self-correction assistance. The generated code is compiled and executed once. If either fails, then the vanilla LLM attempt is marked as unsuccessful. If successful, then a power profile is captured and the code is evaluated by the LLM-as-a-Judge agent (\S\ref{sec:llm-as-a-judge}). Code that passes screening and shows energy savings becomes a candidate for the final solution and seeds the subsequent context-driven refinements in Stages 2 and 3. If the code fails screening or shows no energy savings, then the source code passes to Stage 2.

The same system prompt is used for all inference steps so that differences in results stem from LASSI-EE mechanisms rather than inconsistent prompting.  The prompt frames the LLM as an expert C++/CUDA (or HIP) engineer refactoring code for energy efficiency. 





\subsection{LLM-as-a-Judge: Candidate Code Screening}\label{sec:llm-as-a-judge}

Verifying functional equivalence between generated and source code is a fundamental challenge for automated code generation. LASSI~\cite{LASSI} relied on manual human review, which does not scale. Formal verification methods~\cite{abadi_verifying_2019, siegel_civl_2015, siegel-mironova-avrunin-clarke:2006:symbolic, DBLP:conf/fm/BlomH14} can rigorously prove equivalence under restricted conditions but are computationally expensive, require expert configuration, and do not readily apply to the diverse refactorings produced by LLM-driven generation.

LASSI-EE incorporates an LLM-as-a-Judge agent that provides automated screening of generated code at scale. The Judge receives the source code, the refactored code, and the standard outputs from executing both on identical inputs, along with energy and runtime measurements as contextual information. The Judge evaluates functional equivalence and stdout similarity, returning \texttt{VALID} (passes screening, suitable for energy comparison) or \texttt{INVALID} (filtered from energy metrics). The screening is empirically grounded because it is anchored in executed program behavior, not code inspection alone. For the 18 of 20 HeCBench applications that include built-in correctness validators, the Judge reads PASS/FAIL outcomes embedded in the captured stdout, providing a programmatic signal that supplements the LLM-based comparison.

The Judge prompt instructs the LLM to accept minor output formatting differences (whitespace, layout) and small numerical variations consistent with floating-point reordering that energy-oriented refactorings commonly produce, while flagging algorithmic changes, missing functionality, or substantial output discrepancies as \texttt{INVALID}. While not a substitute for formal verification, the Judge provides practical filtering for large-scale evaluation. Stronger correctness guarantees through formal or differential verification methods remain an important direction for future work.

\subsection{Stage 2: Context Building and Planning}\label{sec:lassiee_2}

Stage 2 builds the structured context that subsequent refactoring stages depend on. The LLM is \emph{self-prompted} to summarize relevant documentation (e.g., CUDA~\cite{nvidia-cuda-guide} or HIP~\cite{amd-hip-guide} programming guide excerpts) and to review the incoming code to describe its functionality and identify inefficiencies. These summaries form the enriched context that informs later stages.

The LLM is then prompted to produce a refactoring plan identifying specific optimization strategies, such as reducing memory transfers, optimizing kernel launch configurations, eliminating redundant computations, improving memory access patterns, or leveraging hardware-specific features. The plan guides Stage 3 refactoring attempts.








\subsection{Stage 3: Iterative Self-Prompted Refactoring}\label{sec:lassiee_3}

This stage is the core iterative refinement loop of LASSI-EE. Using the context and refactoring plan from Stage 2, the LLM repeatedly generates, tests, and evaluates energy-optimized code versions, responding to each prompt with a description of changes and a code block for evaluation.

\textsl{Self-Correcting Feedback Loop.} LASSI-EE inherits the self-correction mechanism from LASSI~\cite{LASSI} for autonomous error recovery. When compilation or runtime errors occur, the pipeline captures the error output and re-prompts the LLM with the buggy code to generate corrections (see Self-correction Loop in Figure \ref{fig:lassiee-pipeline}). This cycle continues until the code compiles and executes without errors, with prompts directing the LLM to preserve function signatures, kernel logic, and original behavior while fixing errors. A configurable maximum threshold prevents runaway iterations. None of the trials reported in \S\ref{sec:results} terminated due to unrepairable code errors. The self-correction mechanism is not applied to the vanilla LLM in Stage 1, ensuring a fair comparison between an unassisted LLM and the full LASSI-EE pipeline.

\textsl{Energy-Driven Iteration and Screening.} Once generated code passes self-correction, the pipeline collects power profiles, calculates energy and runtime metrics, and submits the code to the LLM-as-a-Judge agent (\S\ref{sec:llm-as-a-judge}). If screening fails, the version is archived but filtered from energy comparisons, and Stage 3 restarts with the same refactoring plan. If screening passes, the energy metrics are compared to the current ``best'' code (defined as the lowest-energy candidate among source, vanilla LLM, and prior Stage 3 versions that compiled, executed, and passed screening). If the new version improves on the best, then it becomes the new best, and the LLM is prompted to produce a new refactoring plan reflecting the latest optimizations, initiating another Stage 3 iteration.

\textsl{Stopping Conditions.} Stage 3 terminates after a configurable maximum iteration count, or after a fixed number of consecutive iterations produce no energy improvement. The pipeline tracks consecutive non-improvements through an incrementing parameter that starts at 0.2, increases by 0.2 on each unsuccessful iteration, and resets when an iteration produces a new best version. This parameter doubles as the LLM sampling temperature when supported, broadening response exploration on subsequent iterations.

\begin{figure}[t]
  \centering
  \begin{overpic}[width=1.0\columnwidth]{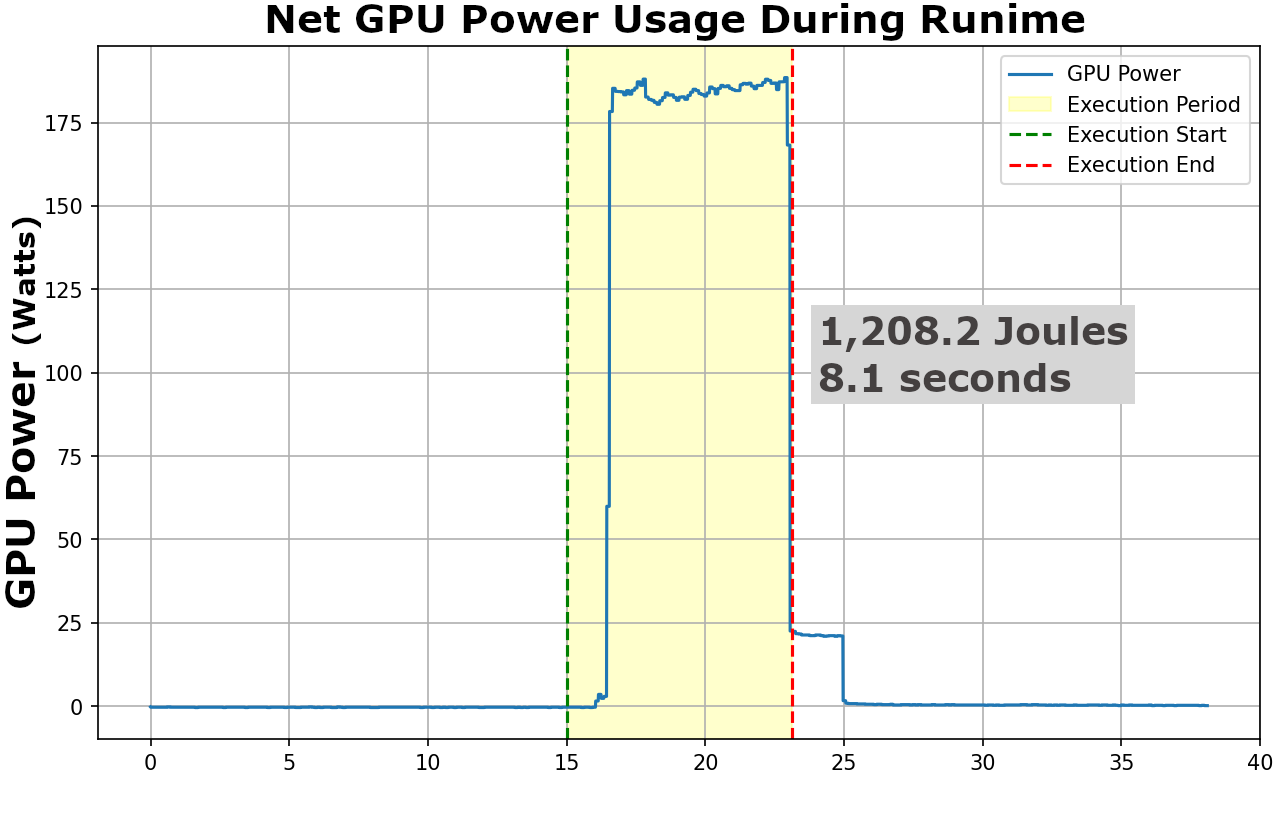}
    \put(10,62){\small\bfseries (a)}
  \end{overpic}
  \begin{overpic}[width=1.0\columnwidth]{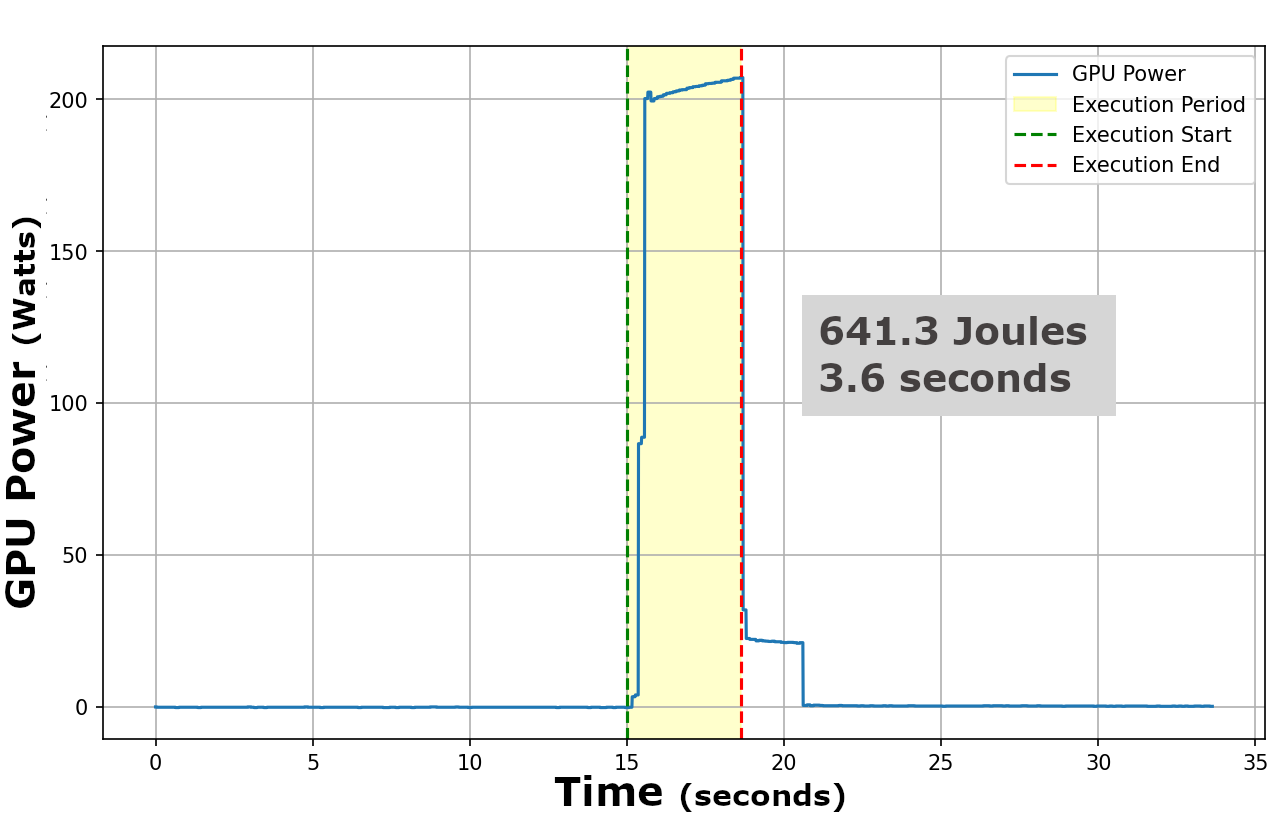}
    \put(10,62){\small\bfseries (b)}
  \end{overpic}
  \caption{Power profiles of Jacobi on an NVIDIA A100 GPU: (a) the original HeCBench implementation~\cite{HeCBench} and (b) the LASSI-EE refactored version. LASSI-EE achieves a 47\% energy reduction through 55\% speedup, despite 19\% increase in average power. Idle power is subtracted.
  }
  \label{fig:jacobi-power-profiles}
\end{figure}

\subsection{Stage 4: Termination and Output}\label{sec:lassiee_4}

Upon termination of Stage 3, the framework identifies the ``best'' code version from all candidates. If no generated code passes screening and reduces energy, the source code is retained as the ``best'' solution, and LASSI-EE reports no energy improvement.



When the ``best'' code achieves energy reduction, such as the \texttt{jacobi} kernel (Figure~\ref{fig:jacobi-power-profiles}), 
LASSI-EE performs a final LLM-based comparison identifying code transformations, energy-relevant optimizations, and potential reasons for the measured energy reduction. This analysis generates a natural language summary documenting the changes. While no automated decisions are based on this comparison, it enables human validation, supports post-hoc analysis, and documents which refactoring tactics yield energy savings on the target platform.


The final ``best'' code is saved along with metadata covering all intermediate code versions, energy profiles, Judge agent verdicts, refactoring plans, iteration counts, and (when applicable) the LLM-generated comparison analysis. This archive supports future LLM fine-tuning for energy-aware code generation and retrospective analysis of the LASSI-EE optimization process.


\section{Experimental Configuration}\label{sec:experiment}

 

\subsection{Applications}

Table~\ref{tab:sourcecode} lists the 22 applications used in this study. Twenty are kernels from the HeCBench suite~\cite{HeCBench}, which provides native CUDA and HIP implementations across diverse computational patterns ready for cross-platform evaluation. Eighteen of these kernels include built-in correctness validation, and the remaining two produce deterministic outputs. To assess longer and more complex codes, we added two miniApps with multiple parallel kernels, miniMDock~\cite{minimdock} and XSBench~\cite{xsbench}. We merged their source files and modified the build system to enable extended codes of 2,394 lines for miniMDock and 1,472 lines for XSBench, exceeding the largest HeCBench application. We verified that all applications executed correctly.

\begin{table}[t]
\caption{Characteristics of the source code applications.}
\label{tab:sourcecode}
\begin{center}
\begin{threeparttable}
    \renewcommand{\arraystretch}{1.2}%
    \resizebox{\columnwidth}{!}{%
\begin{tabular}{@{} p{3cm} l c l }
\toprule
\emph{Category} & \emph{Application} & \textbf{Lines of Code} & \textbf{Input args}  \\ 
\midrule
\rowcolor[HTML]{F2F2F2}\multicolumn{4}{@{}l}{\textbf{HeCBench}}\\
\midrule
{Bandwidth} & randomAccess     & 158   & [10]   \\ 
\midrule
{Bioinformatics} & all-pairs-distance & 328 & [10000] \\
\midrule
\multirow{2}{*}{\makecell[l]{Computer vision\\ and image processing}} 
  & colorwheel       & 154   & [10000, 8, 1] \\
  & marchingCubes*   & 574   & [100] \\
\midrule
{Cryptography} & chacha20* & 130 & [300000] \\
\midrule
\multirow{2}{*}{\makecell[l]{Data compression \\ and reduction}}
 & segment-reduce & 95 & [16384, 100] \\
 & \multicolumn{3}{c}{} \\[1mm]
\midrule
\multirow{3}{*}{\makecell[l]{Data encoding,\\decoding, or \\ verification}} 
  & entropy*       & 171   & [10000, 1024, 1] \\
  & murmurhash3    & 245   & [750000, 500] \\
  & \multicolumn{3}{c}{} \\[1mm]
\midrule    
{Graph and Tree} & floydwarshall & 295 & [2048, 100, 256] \\
\midrule
\multirow{2}{*}{\makecell[l]{Language and kernel \\ features}} 
  & layout         & 197   & [2500] \\
  & threadfence    & 142   & [100, 250000000] \\
\midrule     
{Machine learning}  & dense-embedding   & 193   & [10000, 8, 1]      \\ 
\midrule
\multirow{3}{*}{Math} 
  & jacobi\dag        & 235   & None \\
  & jaccard           & 417   & [1024, 512, 1000] \\
  & matrix-rotate     & 67    & [30000, 100] \\ 
\midrule
\multirow{2}{*}{Search} 
  & bsearch       & 279   & [10000, 1] \\
  & keogh*        & 143   & [256, 22500000, 100] \\
\midrule
{Signal processing} & extrema     & 349   & [750]  \\     
\midrule
\multirow{2}{*}{Simulation} 
  & pathfinder          & 286   & [10000, 1000, 1000] \\
  & lid-driven-cavity   & 1079  & None \\  
\midrule
\rowcolor[HTML]{F2F2F2}\multicolumn{4}{@{}l}{\textbf{miniApps}}\\
\midrule
{Particle transport} & XSBench       & 1472   & [1700000000] \\ 
\midrule
{Molecular dynamics} & miniMDock     & 2394   & [7cpa\_ligand.pdbqt, -nrun 500] \\ 
\bottomrule
\end{tabular}%
}
\begin{tablenotes}[para,flushleft]
    \scriptsize
    * Code includes a dependency that was not considered during refactoring. \\
    \dag Because this app does not accept runtime arguments, we modified the thread \\
    block size multiplier from 2048 to 16384 in the source code. \\
    $\ddagger$ App contains parallel and serial components,  so the serial portion was removed.
\end{tablenotes}
\end{threeparttable}
\end{center}
\end{table}

\subsection{Target Platforms}

Our experiments were conducted on the Chameleon testbeds ~\cite{Chameleon} at UC and TACC. We evaluated LASSI-EE on two GPU platforms: \textbf{NVIDIA A100} (80 GB GPU memory, NVIDIA driver version 560.35.05, CUDA 12.6) and \textbf{AMD MI100} (32 GB GPU memory, ROCm driver version 6.4.1). All 22 applications were executed on both platforms, using CUDA C++ for A100 and HIP C++ for MI100. The LASSI-EE codebase is implemented in Python 3.12.3. During a pipeline run, the framework invokes system-state queries, compilation, and execution directly from the running pipeline process.




\textsl{Code Compilation.} All codes were compiled in the same compute environment with consistent flags across all experimental runs. For CUDA codes on the NVIDIA A100:

{\footnotesize
\begin{verbatim}
nvcc -std=c++14 -Xcompiler -Wall -arch=sm_80 -O3
\end{verbatim}
}

For HIP codes on the AMD MI100:
{\footnotesize
\begin{verbatim}
hipcc -std=c++14 -Wall -O3 --amdgpu-target=gfx908
\end{verbatim}
}




\textsl{Power Measurement.} We focus on GPU power consumption. For the NVIDIA A100, the pipeline utilizes \verb|pyNVML|, the Python bindings to the NVIDIA Management Library, to collect instantaneous power measurements from the GPU. For the AMD MI100, the pipeline uses the \verb|rocm-smi| command-line utility with the \verb|--showpower| flag.


Power usage is sampled every 0.01 seconds on both platforms, beginning with a brief initial period to estimate average idle power. During code execution, power is collected in a separate thread at the same interval. Timestamps are recorded at the launch and completion of code execution to mark the runtime window. Measurements continue for 15 seconds post-execution to capture a stable post-run idle state. Pre- and post-run idle values are averaged to estimate idle system power during execution.

To ensure accurate comparisons, we subtract idle power from all measurements to calculate \emph{net power usage}. This correction accounts for idle power drift, especially during longer sessions. \emph{Net energy} is then estimated by summing $N$ idle-subtracted power samples $P_i$ multiplied by the time interval of 0.01 s. Any negative samples from measurement noise are clamped to zero to avoid non-physical contributions to the energy total.

\subsection{LLM Models}

LASSI-EE is designed as an LLM-agnostic pipeline. In this study, we employed a multi-model approach to leverage strengths of different LLMs while maintaining consistency across our experimental trials. We used o4-mini for context building and code generation across the 20 HeCBench applications. For the two miniApps, we used GPT-5 as the code generator due to context length constraints in o4-mini, while retaining o4-mini for context building. We used GPT-4.1 as the LLM-as-a-Judge agent for its larger context window and detailed evaluation capabilities. Neither o4-mini nor GPT-5 supports temperature tuning. LASSI-EE's iterative loop uses an incrementing parameter both to broaden exploration through the LLM sampling temperature and to track consecutive non-improving iterations toward Stage 3 termination (§II-E). For the experiments reported here, the iteration tracking was active but the temperature value had no effect on inference.


We set $n=30$ independent generation attempts per application per device, providing the basis for the energy analysis in §IV. Each run executes both a vanilla LLM generation (§II-B) and the complete LASSI-EE pipeline, producing paired results. A generated code sample passes screening if it (i) compiles, (ii) executes without errors, and (iii) receives a \texttt{VALID} verdict from the LLM-as-a-Judge.


\section{Experimental Results}\label{sec:exp-results}
\subsection{Overall Results}\label{sec:results}
 

Our experiments comprise 1,320 trials (22 applications × 2 platforms × 30 trials per application-platform pair).
Table~\ref{tab:performance_summary} presents measured performance for the applications on both AMD MI100 and NVIDIA A100 GPUs. For each application, the consumed energy (J), average instantaneous power (W), and runtime (s) for the source code are averaged over 30 independent trials. The corresponding LASSI-EE measurements are averaged over only those trials that passed the LLM-as-a-Judge screening and consumed less total energy than the source code, with differences shown in parentheses. The `Best Trial' columns report pipeline metadata for the lowest-energy code generation attempt for the application/GPU pair. `Self-Corrections' indicates iterations required to fix compilation or execution errors (see the \emph{Self-correction Loop} in Figure~\ref{fig:lassiee-pipeline}), and `Iterations for EE' is the iteration number where the most energy-efficient code was generated. For example, the best \texttt{segment-reduce} refactoring on MI100 had no bugs and was generated during the fourth refactoring attempt.

\textsl{Pipeline Cost.} Average pipeline runtime per trial is approximately 30 minutes, with a range from about 10 minutes to outliers of 80--114 minutes (Table~\ref{tab:performance_summary}). Two factors drive trial duration. Compilation or execution failures trigger self-correction iterations that add overhead. Stage 3 also continues as long as the pipeline keeps finding improvements, with each new improvement resetting the counter of consecutive non-improving iterations that gates termination (§\ref{sec:lassiee_3}). Trials in which the pipeline finds gains late in Stage 3 therefore terminate later than trials that plateau early.


\FloatBarrier
\begin{table*}[!htbp]
\centering
\caption{Performance characteristics comparing source code and LASSI-EE-generated code for representative applications. Energy, power, and runtime show LASSI-EE performance with percentage change ($\Delta$\%) in parentheses. Negative values indicate reductions (improvements). These metrics represent conditional averages over trials that compiled, executed successfully, passed LLM-as-a-Judge screening, and consumed less energy than source code. Best-trial statistics report values from the trial with lowest energy per application/GPU. Pipeline time shows average across all successful trials. Cross-app averages are computed over all 22 applications. The complete results for all applications are available in our open-source GitHub repository. 
}
\label{tab:performance_summary}
\resizebox{\textwidth}{!}{%
\begin{tabular}{@{}l >{\centering\arraybackslash}p{0.7cm} cc cc cc cc c@{}}
\toprule
\multirow{2}{*}{\textbf{Application}} & \multirow{2}{*}{\textbf{GPU}} & \multicolumn{2}{c}{\textbf{Energy (J)}} & \multicolumn{2}{c}{\textbf{Power (W)}} & \multicolumn{2}{c}{\textbf{Runtime (s)}} & \multicolumn{2}{c}{\textbf{Best Trial}} & \textbf{Avg Pipeline} \\
\cmidrule(lr){3-4} \cmidrule(lr){5-6} \cmidrule(lr){7-8} \cmidrule(lr){9-10}
 & & \textbf{Source} & \textbf{LASSI-EE ($\Delta$\%)} & \textbf{Source} & \textbf{LASSI-EE ($\Delta$\%)} & \textbf{Source} & \textbf{LASSI-EE ($\Delta$\%)} & \textbf{Self-Corr.} & \textbf{Iter. EE} & \textbf{Time (min)} \\
\midrule

\multirow{2}{*}{segment-reduce} & MI100 & 710.9 & 468.0 ($-$34.2) & 62.5 & 128.0 (+104.6) & 11.5 & 4.1 ($-$64.6) & 0 & 4 & 28.9 \\
 & A100 & 622.0 & 289.8 ($-$53.4) & 38.8 & 79.3 (+104.3) & 16.0 & 3.5 ($-$78.1) & 0 & 5 & 13.4 \\
\cmidrule(lr){1-11}

\multirow{2}{*}{matrix-rotate} & MI100 & 2591.8 & 815.0 ($-$68.6) & 75.9 & 67.6 ($-$10.9) & 34.3 & 10.3 ($-$70.0) & 0 & 1 & 34.7 \\
& A100 & 850.1 & 604.1 ($-$28.9) & 21.9 & 17.6 ($-$19.7) & 38.8 & 35.5 ($-$8.4) & 0 & 1 & 19.7 \\
\cmidrule(lr){1-11}

\multirow{2}{*}{chacha20} & MI100 & 1008.5 & 775.5 ($-$23.1) & 38.8 & 36.3 ($-$6.2) & 26.0 & 21.0 ($-$19.2) & 0 & 10 & 28.1 \\
& A100 & 282.0 & 237.3 ($-$15.9) & 26.8 & 25.6 ($-$4.7) & 10.5 & 9.0 ($-$14.7) & 0 & 10 & 17.9 \\
\cmidrule(lr){1-11}

\multirow{2}{*}{threadfence} & MI100 & 194.1 & 40.5 ($-$79.1) & 56.4 & 27.1 ($-$51.9) & 3.4 & 1.5 ($-$56.7) & 0 & 6 & 27.4 \\
& A100 & 365.7 & 24.7 ($-$93.3) & 50.8 & 14.6 ($-$71.4) & 7.2 & 1.3 ($-$81.8) & 0 & 7 & 18.2 \\
\cmidrule(lr){1-11}

\multirow{2}{*}{keogh} & MI100 & 377.2 & 267.0 ($-$29.2) & 29.5 & 21.3 ($-$27.8) & 12.8 & 12.4 ($-$3.2) & 0 & 4 & 18.8 \\
& A100 & 391.1 & 356.3 ($-$8.9) & 8.6 & 8.0 ($-$7.1) & 44.6 & 44.8 (+0.5) & 0 & 3 & 17.0 \\
\cmidrule(lr){1-11}

\multirow{2}{*}{colorwheel} & MI100 & 11873.4 & 576.7 ($-$95.1) & 76.6 & 86.7 (+13.2) & 155.1 & 7.0 ($-$95.5) & 1 & 7 & 32.7 \\
& A100 & 2309.9 & 79.0 ($-$96.6) & 103.0 & 37.8 ($-$63.3) & 22.4 & 1.7 ($-$92.6) & 0 & 2 & 16.5 \\
\cmidrule(lr){1-11}

\multirow{2}{*}{randomAccess} & MI100 & 709.4 & 166.7 ($-$76.5) & 52.1 & 37.6 ($-$27.8) & 13.6 & 4.4 ($-$67.3) & 0 & 3 & 25.4 \\
& A100 & 359.8 & 176.2 ($-$51.0) & 21.6 & 21.5 ($-$0.6) & 16.6 & 9.4 ($-$43.5) & 0 & 11 & 20.2 \\
\cmidrule(lr){1-11}

\multirow{2}{*}{entropy} & MI100 & 1.6 & 0.3 ($-$82.7) & 0.9 & 0.2 ($-$83.3) & 1.8 & 1.8 (0) & 0 & 9 & 19.6 \\
& A100 & 1718.2 & 1659.1 ($-$3.4) & 55.8 & 58.1 (+4.0) & 31.3 & 31.1 ($-$0.6) & 0 & 5 & 16.3 \\
\cmidrule(lr){1-11}

\multirow{2}{*}{dense-embedding} & MI100 & 453.8 & 336.7 ($-$25.8) & 3.2 & 2.5 ($-$23.4) & 140.2 & 134.1 ($-$4.3) & 0 & 11 & 37.8 \\
& A100 & 636.7 & 377.3 ($-$40.7) & 16.9 & 13.4 ($-$20.6) & 37.6 & 27.4 ($-$27.1) & 0 & 8 & 19.8 \\
\cmidrule(lr){1-11}

\multirow{2}{*}{layout} & MI100 & 168.4 & 143.4 ($-$14.8) & 73.4 & 71.3 ($-$2.9) & 2.3 & 2.0 ($-$10.7) & 0 & 2 & 20.8 \\
& A100 & 1284.0 & 966.4 ($-$24.7) & 66.4 & 74.4 (+12.1) & 19.3 & 12.8 ($-$33.5) & 0 & 2 & 13.6 \\
\cmidrule(lr){1-11}

\multirow{2}{*}{jacobi} & MI100 & 2200.9 & 1597.2 ($-$27.4) & 125.1 & 156.6 (+25.2) & 17.8 & 10.4 ($-$41.6) & 0 & 1 & 82.2 \\
& A100 & 1216.3 & 648.1 ($-$46.7) & 150.9 & 131.7 ($-$12.8) & 8.1 & 4.7 ($-$42.3) & 0 & 5 & 22.2 \\
\cmidrule(lr){1-11}

\multirow{2}{*}{murmurhash3} & MI100 & 1144.9 & 1061.8 ($-$7.3) & 94.9 & 108.8 (+14.7) & 12.2 & 10.1 ($-$17.2) & 0 & 4 & 26.7 \\
& A100 & 1452.7 & 1304.4 ($-$10.2) & 69.4 & 65.8 ($-$5.1) & 20.9 & 19.7 ($-$5.8) & 0 & 3 & 18.4 \\
\cmidrule(lr){1-11}

\multirow{2}{*}{bsearch} & MI100 & 6996.5 & 6599.1 ($-$5.7) & 222.8 & 214.0 ($-$3.9) & 31.4 & 29.8 ($-$5.0) & 0 & 8 & 114.7 \\
& A100 & 2259.5 & 1845.0 ($-$18.3) & 198.6 & 177.4 ($-$10.7) & 11.4 & 9.5 ($-$16.2) & 0 & 5 & 13.1 \\
\cmidrule(lr){1-11}

\multirow{2}{*}{pathfinder} & MI100 & 1222.9 & 208.4 ($-$83.0) & 55.2 & 11.2 ($-$79.7) & 22.2 & 11.0 ($-$50.6) & 0 & 11 & 25.3 \\
& A100 & 1933.0 & 181.0 ($-$90.6) & 56.4 & 6.0 ($-$89.4) & 34.4 & 22.9 ($-$33.5) & 0 & 7 & 16.6 \\
\cmidrule(lr){1-11}

\multirow{2}{*}{floydwarshall} & MI100 & 1388.1 & 1206.9 ($-$13.1) & 124.0 & 112.7 ($-$9.1) & 11.2 & 10.5 ($-$6.5) & 1 & 9 & 21.8 \\
& A100 & 817.0 & 752.7 ($-$7.9) & 51.3 & 48.2 ($-$6.0) & 15.9 & 15.5 ($-$2.9) & 0 & 10 & 15.8 \\
\cmidrule(lr){1-11}

\multirow{2}{*}{all-pairs-distance} & MI100 & 2922.1 & 1526.6 ($-$47.8) & 133.4 & 167.7 (+25.7) & 21.9 & 10.0 ($-$54.4) & 1 & 5 & 35.9 \\
& A100 & 836.6 & 639.4 ($-$23.6) & 81.0 & 93.8 (+15.8) & 10.5 & 6.7 ($-$36.2) & 0 & 4 & 20.3 \\
\cmidrule(lr){1-11}

\multirow{2}{*}{extrema} & MI100 & 2580.2 & 2399.1 ($-$7.0) & 155.0 & 146.5 ($-$5.5) & 16.7 & 16.4 ($-$1.6) & 0 & 6 & 37.5 \\
& A100 & 2693.0 & 1798.5 ($-$33.2) & 30.9 & 29.9 ($-$3.1) & 91.5 & 100.7 (+10.1) & 0 & 1 & 27.8 \\
\cmidrule(lr){1-11}

\multirow{2}{*}{jaccard} & MI100 & 4755.4 & 3395.2 ($-$28.6) & 103.3 & 95.9 ($-$7.2) & 46.0 & 34.9 ($-$24.1) & 0 & 5 & 45.8 \\
& A100 & 1218.6 & 1028.5 ($-$15.6) & 79.8 & 71.0 ($-$11.0) & 15.3 & 14.3 ($-$6.4) & 0 & 5 & 24.2 \\
\cmidrule(lr){1-11}

\multirow{2}{*}{marchingCubes} & MI100 & 1201.8 & 1171.5 ($-$2.5) & 83.6 & 81.9 ($-$2.0) & 14.4 & 14.3 ($-$0.5) & 0 & 6 & 31.2 \\
& A100 & 438.0 & 398.5 ($-$9.0) & 36.7 & 35.9 ($-$2.3) & 11.9 & 11.6 ($-$2.6) & 0 & 4 & 25.5 \\
\cmidrule(lr){1-11}

\multirow{2}{*}{lid-driven-cavity} & MI100 & 2358.1 & 2119.8 ($-$10.1) & 122.1 & 127.8 (+4.6) & 19.3 & 16.6 ($-$14.0) & 0 & 9 & 43.9 \\
& A100 & 1557.1 & 1192.0 ($-$23.5) & 36.2 & 37.6 (+3.9) & 43.5 & 36.7 ($-$15.7) & 0 & 7 & 42.8 \\
\cmidrule(lr){1-11}

\multirow{2}{*}{XSBench} & MI100 & 5864.5 & 5531.6 ($-$5.7) & 142.9 & 168.6 (+18.0) & 41.7 & 33.1 ($-$20.7) & 0 & 3 & 53.8 \\
& A100 & 3038.4 & 2753.7 ($-$9.4) & 155.9 & 141.1 ($-$9.5) & 19.5 & 20.7 (+5.9) & 0 & 6 & 37.9 \\
\cmidrule(lr){1-11}

\multirow{2}{*}{miniMDock} & MI100 & 5059.8 & 3625.8 ($-$28.3) & 223.3 & 205.6 ($-$7.9) & 22.7 & 17.5 ($-$22.9) & 0 & 4 & 50.2 \\
& A100 & 2627.2 & 1568.3 ($-$40.3) & 133.2 & 163.7 (+23.0) & 19.7 & 9.5 ($-$51.9) & 0 & 8 & 47.6 \\

\midrule
\rowcolor{gray!20}
\multicolumn{5}{@{}>{\cellcolor{gray!20}}l}{\textbf{Cross-app average (all 22 apps, $\Delta$\%)}} & {} & {} & {} & {} & {} & {} \\
\rowcolor{gray!20}
{} & MI100 & {} & \textcolor{blue}{($-$36.2\%)} & {} & \textcolor{blue}{($-$6.5\%)} & {} & \textcolor{blue}{($-$29.6\%)} & {} & 5.8 & 38.3 \\
\rowcolor{gray!20}
{} & A100 & {} & \textcolor{blue}{($-$33.9\%)} & {} & \textcolor{blue}{($-$7.9\%)} & {} & \textcolor{blue}{($-$26.2\%)} & {} & 5.4 & 22.0 \\
\bottomrule
\end{tabular}}
\end{table*}


\textsl{Cross-Device Consistency.} When LASSI-EE generated energy-efficient code, it achieved average energy savings of 33.9\% on NVIDIA A100 and 36.2\% on AMD MI100, a difference of only 2.3 percentage points (pp). Average power reductions were 7.9\% and 6.5\%, respectively, while runtime improvements averaged 26.2\% and 29.6\%. This consistency demonstrates that LASSI-EE's energy-aware refactoring generalizes effectively across hardware architectures. Averaged across all 22 applications and both platforms, LASSI-EE achieves 35.1\% energy reduction, 7.2\% power reduction, and 27.9\% runtime reduction.


\textsl{Runtime and Power Trade-offs.} Application-specific energy reductions covered a broad range from 2.5\% (\texttt{marchingCubes} on MI100) to 96.6\% (\texttt{colorwheel} on A100). Nineteen of 22 applications on A100 and 21 of 22 on MI100 showed runtime improvements, with \texttt{entropy} on MI100 essentially unchanged. Several applications showed increased average power and still achieved energy reduction through faster runtimes. For example, \texttt{segment-reduce} had 104.3\% average power increase on A100 and 104.6\% on MI100, but still achieved 53.4\% and 34.2\% energy reductions, respectively, through 78.1\% and 64.6\% speedups. Figure~\ref{fig:jacobi-power-profiles} illustrates this trade-off for \texttt{jacobi}, showing power profiles before and after LASSI-EE refactoring, where 19\% higher instantaneous power accompanied a 55\% speedup to achieve 47\% net energy reduction. Across applications, energy reductions arise from various combinations of power and runtime changes. 


\textsl{Application Code Length.} For the two miniApps, \texttt{XSBench} (1,472 lines) and \texttt{miniMDock} (2,394 lines), LASSI-EE delivered average energy savings of 5.7\% (MI100) to 9.4\% (A100) for XSBench, and 28.3\% (MI100) to 40.3\% (A100) for miniMDock. These codes are approximately 3× longer than most HeCBench codes, indicating LASSI-EE applies to parallel codes well beyond typical benchmark kernel scale.



\subsection{Energy Savings Analysis}\label{sec:energy_savings_analysis}
\begin{figure}[bh!]
  \centering
  \begin{overpic}[width=1.0\columnwidth]{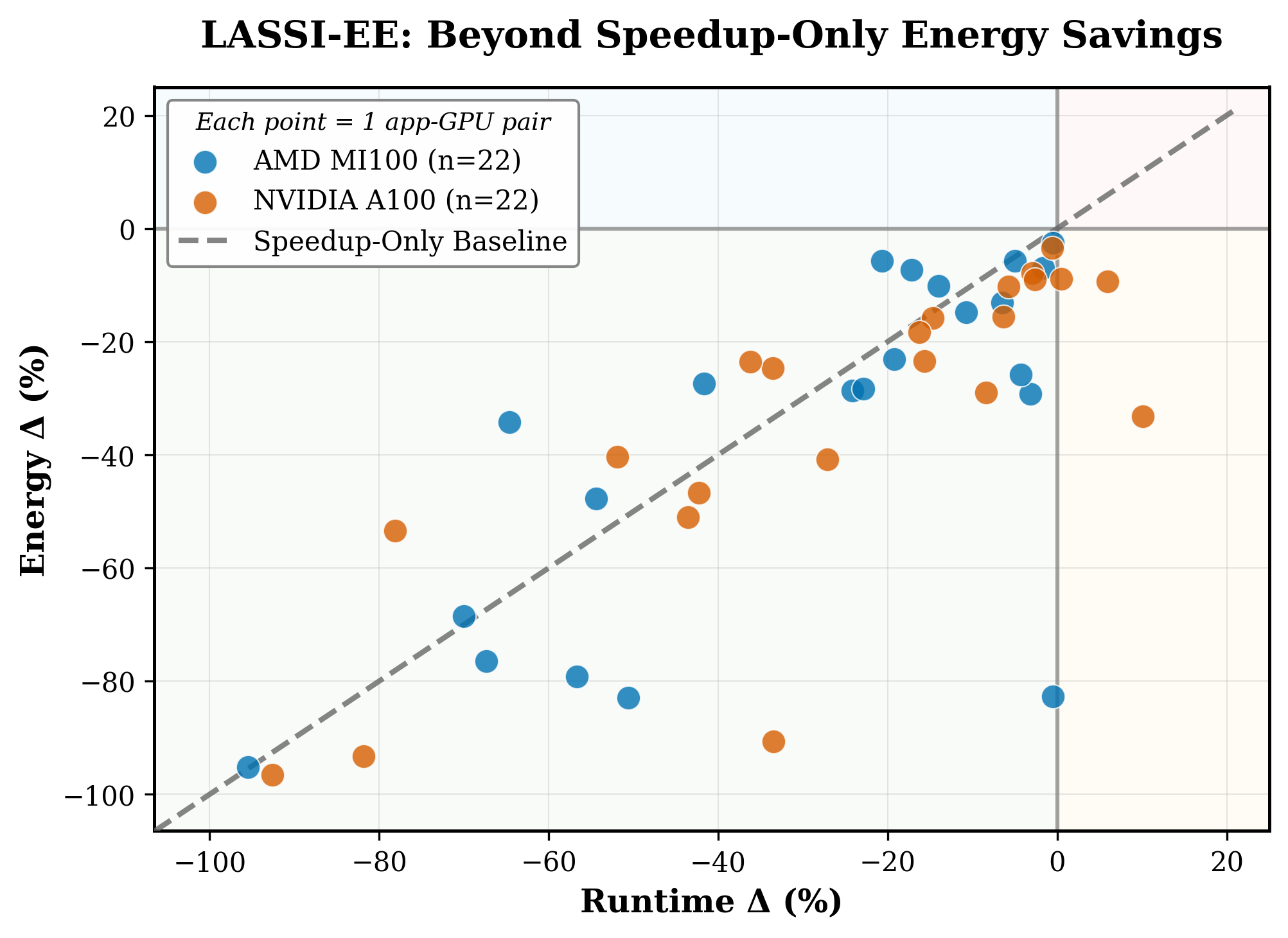}
    \put(2,2){\small\bfseries (a)}
  \end{overpic}
  \begin{overpic}[width=1.0\columnwidth]{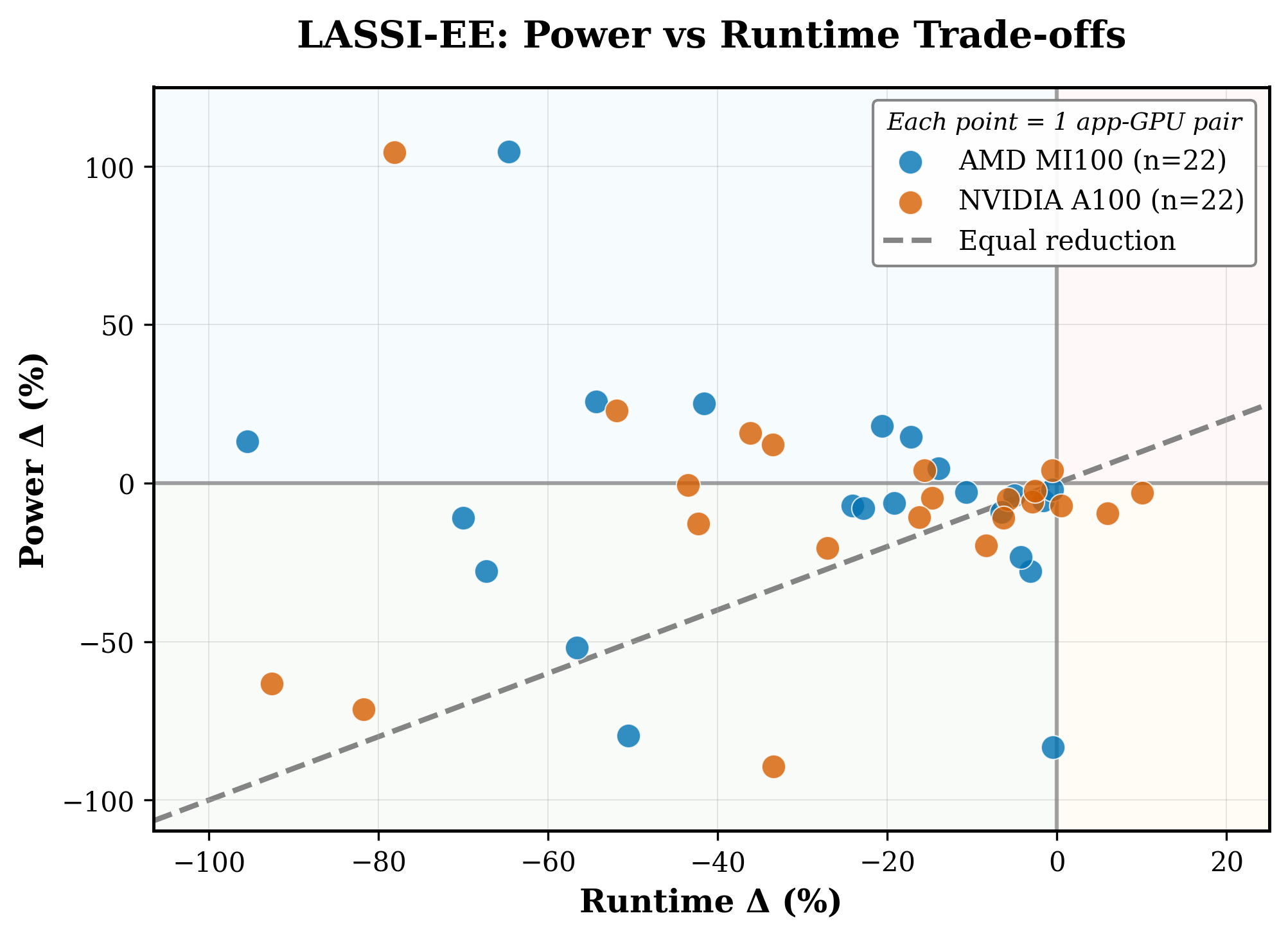}
    \put(2,2){\small\bfseries (b)}
  \end{overpic}
  \caption{Decomposition of LASSI-EE energy savings across all 44 application-device pairs. (a) Energy reduction versus runtime reduction; the diagonal represents a ``speedup-only baseline.'' Points below (72.7\%) indicate power reduction contributed beyond speedup. (b) Power versus runtime change; 63.6\% reduced both, while 29.5\% traded higher power for speedups.}
  \label{fig:energy-power-tradeoffs}
\end{figure}

Because energy equals power multiplied by time, energy reduction can arise from runtime reduction (speedup), power reduction, or both. Figure~\ref{fig:energy-power-tradeoffs} (a) plots energy reduction against runtime reduction for all 44 application-device pairs, with the diagonal representing a ``speedup-only baseline'' for the expected energy savings if power remained constant. Points below this diagonal indicate that power reduction contributed \textit{additional savings} beyond speedup alone (because both axes represent reductions as negative values, points \textit{further below} the diagonal achieved greater energy savings than runtime improvement alone would deliver). Each data point represents the average over trials that passed the screening and achieved energy reduction.

Of the 44 application-device pairs, 72.7\% (32/44) fall below the diagonal, indicating that LASSI-EE achieves energy savings beyond what speedup alone provides. Only 27.3\% of optimizations lie on or above the diagonal, where speedup fully accounts for energy gains. Considering \textit{runtime outcomes}, 93.2\% of pairs achieved both energy and runtime reduction, while 6.8\% reduced energy through pure power optimization despite slight runtime increases. No application-platform pair showed degradation in both metrics.

Figure~\ref{fig:energy-power-tradeoffs} (b) further examines the underlying mechanisms by plotting power change against runtime change. Considering \textit{power outcomes}, 70.4\% achieved power reduction: 63.6\% reduced both power and runtime, while 6.8\% reduced power despite slight runtime increases. This 6.8\% directly evidences energy-aware, rather than runtime-only, optimization, as a runtime-only objective cannot justify any slowdown. Together with the 29.5\% of pairs that trade power for substantial speedups, these results show LASSI-EE adapts its energy-reduction strategy per application rather than optimizing runtime or power in isolation.


Several cases show distinct optimization strategies. The \texttt{entropy} application on MI100 achieved 83.3\% power reduction with essentially unchanged runtime, representing near-pure power optimization. On A100, \texttt{pathfinder} reduced power by 89.4\%, exceeding its 33.5\% runtime improvement. Conversely, \texttt{segment-reduce} exemplifies speedup-driven optimization: 78\% runtime reduction on A100 outweighed a 104\% power increase to achieve 53\% energy savings.


\subsection{Optimization Techniques for Energy Savings}\label{sec:opt-techniques}

We characterize LASSI-EE's optimization strategies through a broad categorization of over 4,000 optimization methods extracted across all successful code generations. These optimization methods were automatically identified during Stage 4 termination (\S\ref{sec:lassiee_4}), where the LLM generated a comparison report between the original source code and the final refactored version, explicitly listing the optimizations applied.

\begin{figure}[t]
  \centering
  \begin{overpic}[width=0.48\columnwidth]{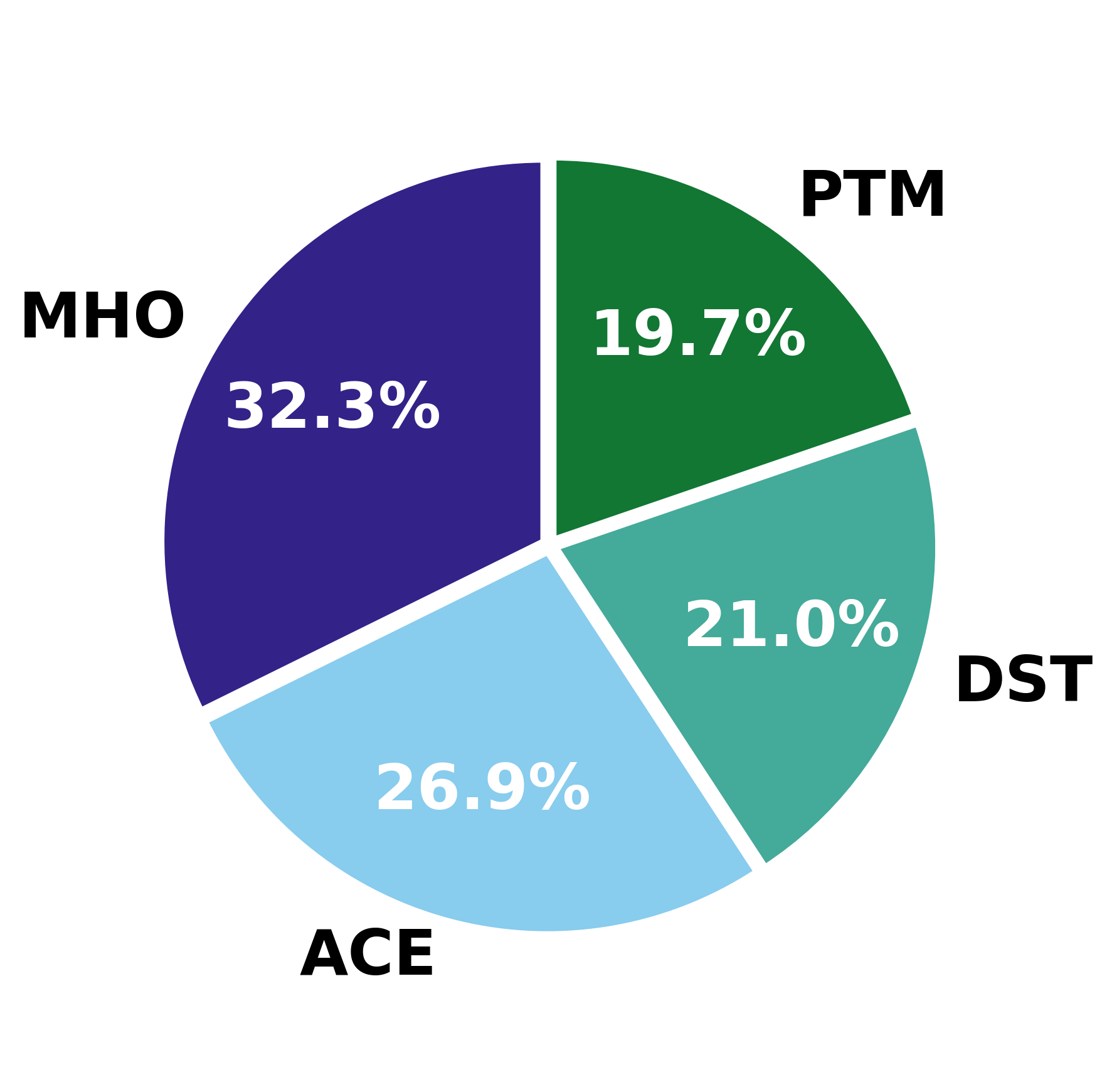}
    \put(1,96){\small\bfseries (a)}
  \end{overpic}
  \hfill 
  \begin{overpic}[width=0.48\columnwidth]{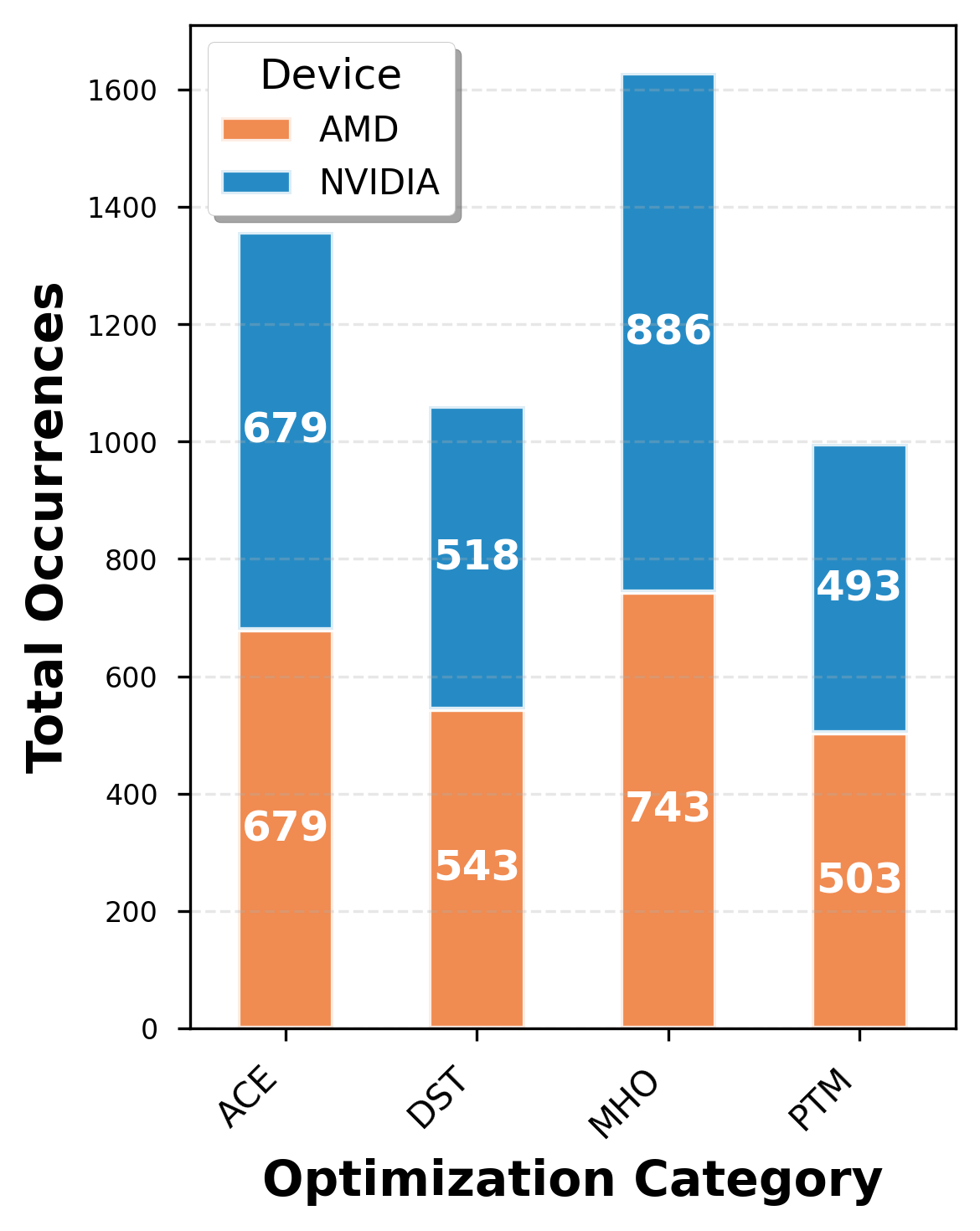}
    \put(1,96){\small\bfseries (b)}
  \end{overpic}
  \caption{Optimization category distribution across all applications: (a) Overall distribution shows balanced utilization across four categories: Memory Hierarchy Optimization (MHO), Algorithmic \& Computational Efficiency (ACE), Device-Specific Tuning (DST), and Parallelism \& Thread Management (PTM). (b) Cross-device comparison show consistent categorical emphasis on both AMD MI100 and NVIDIA A100.}
  \label{fig:opt-categories}
\end{figure}

We sorted these optimizations into four high-level categories based on their optimization goal: Memory Hierarchy Optimization (MHO), Algorithmic \& Computational Efficiency (ACE), Device-Specific Tuning (DST), and Parallelism \& Thread Management (PTM). The relatively balanced distribution across these categories (Figure \ref{fig:opt-categories}a) demonstrates that LASSI-EE employs diverse optimization strategies rather than over-relying on any single approach. 

This balanced approach is consistent across both AMD MI100 and NVIDIA A100 platforms (Figure \ref{fig:opt-categories}b), with each device showing similar proportional emphasis on all four category types. 

The near-pure power optimization in \texttt{entropy} on MI100 was driven primarily by MHO techniques: placement of lookup tables in \texttt{\_\_constant\_\_} memory appeared in 87\% of successful trials, complemented by shared-memory tiling and per-thread register histograms to minimize global memory traffic. The combined power-and-runtime reduction in \texttt{pathfinder} on A100 leveraged both MHO and PTM optimizations, with pinned host memory and \texttt{\_\_ldg()} read-only caching appearing alongside warp-aligned block sizing and occupancy-based configuration. The speedup-driven optimization in \texttt{segment-reduce} relied heavily on ACE techniques: on-the-fly key generation with Thrust iterators and pre-allocated buffer reuse eliminated redundant memory operations, dramatically increasing GPU utilization (and instantaneous power) while reducing runtime by 78\%.

For \texttt{keogh} and \texttt{entropy} on A100, where vanilla LLM achieved zero energy reduction, LASSI-EE's successful refactorings employed both ACE and MHO techniques. For \texttt{keogh}, ACE optimizations dominated, including precomputing reciprocals to replace divisions and hoisting loop-invariant index arithmetic. For \texttt{entropy}, MHO techniques drove improvements through constant-memory placement of lookup tables and pinned host memory, complemented by loop unrolling.

Notably, none of the identified optimizations introduced mixed-precision or precision-reducing transformations. LASSI-EE achieved its energy savings through memory hierarchy, algorithmic, and parallelism optimizations rather than precision trade-offs that could compromise numerical accuracy in scientific applications.

We also observe strong device specialization. Of the approximately 3,500 \textit{unique} optimization implementations identified across both platforms, only 124 (3.5\%) were \emph{syntactically identical} in the generated codes for AMD MI100 and NVIDIA A100, while 47\% were AMD-specific and 49\% were NVIDIA-specific. This distribution suggests LASSI-EE appropriately translates optimization \emph{concepts} into platform-specific APIs. For example, pinned host memory allocation appears frequently in both AMD and NVIDIA code but uses \texttt{hipHostMalloc} on AMD versus \texttt{cudaHostAlloc} on NVIDIA. Similarly, read-only data caching is implemented via compiler optimization hints on AMD but leverages NVIDIA's hardware-specific \texttt{\_\_ldg} intrinsic on A100 GPUs. This pattern demonstrates that LASSI-EE adapts conceptual optimization strategies to device-appropriate implementations.

The most frequently applied optimizations across both platforms included variations of pinned (page-locked) host memory allocation~\cite{CUDApinnedmem} (MHO), \texttt{\_\_restrict\_\_} pointer annotations (ACE) for aliasing disambiguation, and constant memory placement (MHO) for frequently accessed read-only data. 
Differences in optimization frequency were observed between devices: generated HIP codes for AMD showed higher utilization of precomputed lookup tables in constant memory (22.7\% mean appearance rate), asynchronous device-to-host transfers (14.2\%), and HIP-specific event timing with \texttt{hipEvent} (8.2\%)~\cite{amd-hip-guide}, while CUDA code for NVIDIA more frequently employed the platform-exclusive \texttt{\_\_ldg} read-only cache intrinsic (16.0\%)~\cite{nvidia-cuda-guide}, loop unrolling pragmas (15.9\%), and CUDA-specific pinned memory allocation functions (14.0\%). 

\emph{Overall, the analysis demonstrates LASSI-EE’s ability to select effective optimizations during refactoring, combining general techniques with platform-specific APIs to produce energy-efficient code variants.}

\section{Energy Reduction under LLM Non-Determinism}\label{sec:energy-reduction-at-k}

The results in \S\ref{sec:exp-results} characterize LASSI-EE on trials that produced passing energy-reducing code refactorings across 30 runs. In practice, LLM inference produces a distribution of outputs across runs, and not every run yields the same outcome. This section quantifies expected energy savings across multi-attempt generations and identifies a favorable operating point (i.e., favorable number of runs) for practical deployment.

\subsection{The energy-reduction@$k$ Metric}\label{sec:metrics}

Inspired by the pass@$k$ metric~\cite{chenevaluating2021}, we introduce \textbf{energy-reduction@$k$}, a novel metric for energy-aware code generation that quantifies expected optimization quality: ``If I generate $k$ code candidates and select the most energy-efficient one that passes screening, what energy reduction can I expect?'' Formally, let $\bar{E}_{\text{source}}$ denote the average energy consumption of the source code across all trial executions, and let $E_i$ represent the energy consumption of candidate $i$ from the subset of screened generated codes. We define:
\begin{multline}
\label{eq:er_at_k}
\textbf{energy-reduction@}k := \mathbb{E}\!\left[1 - \frac{\binom{n-c}{k}}{\binom{n}{k}}\right] \times \\
\mathbb{E}\!\left[\max\!\left\{\!\left(\frac{\bar{E}_{\text{source}} - E_i}{\bar{E}_{\text{source}}}\right)^{\!+} \!\!\mid i \in c_k\right\}\right]
\end{multline}
where $c$ denotes the number of codes that pass candidate screening; $c_k$ represents $k$ candidates sampled from generated codes that passed screening among $n$ independent runs; $E_i$ denotes the energy measurement of candidate $i$ that received a \texttt{VALID} screening verdict; and $\mathbb{E}[\cdot]$ denotes expectation. The ratio represents the relative energy reduction achieved by candidate $i$ compared to the source code, and the first factor on the right-hand side is pass@$k$, which is independent of the energy reduction factor.

\subsection{The energy-reduction@$k$ Results}
\label{sec:energy_reduction_results}

\begin{table}[h]
\centering
\caption{\textbf{energy-reduction@$k$ Snapshot}: Expected energy reduction for representative applications. All values as percentages show the expected \emph{energy savings} over the source code in the LASSI-EE-generated parallel code, with differences over vanilla LLM in parentheses ($\Delta$ pp). The cross-application results are averaged over all 22 benchmark source codes, showing similar performance on A100 and MI100 for each $k$ on average. The complete results for all applications are available in our open-source GitHub repository. 
}
\footnotesize
\begin{tabular}{@{}lccc@{}}
\toprule
\multirow{2}{*}{\textbf{App/GPU}} & \textbf{$k=1$} & \textbf{$k=3$} & \textbf{$k=5$} \\
 & \textbf{LASSI-EE ($\Delta$)} & \textbf{LASSI-EE ($\Delta$)} & \textbf{LASSI-EE ($\Delta$)} \\
\midrule
\rowcolor{gray!20}
\multicolumn{3}{@{}l}{\textbf{Cross-app average} (all 22 apps)} & {} \\
\rowcolor{gray!20}
\quad MI100 & 29.6\% \textcolor{blue}{(+19.1)} & 49.5\% \textcolor{blue}{(+22.5)} & 55.2\% \textcolor{blue}{(+19.2)} \\
\rowcolor{gray!20}
\quad A100 & 28.6\% \textcolor{blue}{(+18.0)} & 46.9\% \textcolor{blue}{(+21.4)} & 54.1\% \textcolor{blue}{(+20.9)} \\
\rowcolor{gray!20}
\textbf{Cross-device avg:} & 29.1\% & 48.2\% & 54.6\% \\
\midrule
\textbf{segment-reduce} & & & \\
\quad MI100 & 28.5 (+18.9) & 57.4 (+33.4) & 68.7 (+38.6) \\
\quad A100 & 48.1 (+25.0) & 66.1 (+24.8) & 74.0 (+26.5) \\
\cmidrule(lr){1-4}
\textbf{jacobi} & & & \\
\quad MI100 & 22.7 (+20.5) & 46.5 (+36.3) & 52.5 (+34.6) \\
\quad A100 & 44.8 (+38.8) & 66.4 (+37.7) & 75.2 (+28.3) \\
\cmidrule(lr){1-4}
\textbf{entropy} & & & \\
\quad MI100 & 63.9 (+54.9) & 88.0 (+52.1) & 90.7 (+39.0) \\
\quad A100 & 3.2 (+3.2) & 6.5 (+6.5) & 8.8 (+8.8) \\
\cmidrule(lr){1-4}
\textbf{keogh} & & & \\
\quad MI100 & 13.7 (+11.4) & 46.4 (+36.3) & 59.1 (+41.6) \\
\quad A100 & 1.0 (+1.0) & 5.9 (+5.9) & 11.7 (+11.7) \\
\cmidrule(lr){1-4}
\textbf{lid-driven-cavity} & & & \\
\quad MI100 & 7.8 (+7.7) & 16.4 (+15.9) & 22.9 (+21.9) \\
\quad A100 & 16.6 (+13.9) & 41.3 (+24.7) & 51.0 (+17.7) \\
\cmidrule(lr){1-4}
\textbf{XSBench} & & & \\
\quad MI100 & 3.0 (+0.7) & 10.6 \textcolor{red}{($-$2.4)} & 16.3 \textcolor{red}{($-$4.9)} \\
\quad A100 & 5.3 (+5.2) & 15.8 (+15.2) & 19.4 (+18.2) \\
\cmidrule(lr){1-4}
\textbf{miniMDock} & & & \\
\quad MI100 & 20.8 (+9.5) & 44.1 (+24.8) & 51.1 (+25.7) \\
\quad A100 & 29.6 (+15.1) & 50.8 (+18.8) & 57.9 (+17.4) \\
\bottomrule
\end{tabular}
\label{tab:energy_reduction_summary}
\end{table}

Table~\ref{tab:energy_reduction_summary} lists a snapshot of the expected energy-reduction@$k$ metrics for vanilla LLM and LASSI-EE across representative applications on both GPU platforms. 
This metric combines the probability of \texttt{VALID} code screening by the LLM-as-a-Judge (pass@$k$) with measured energy savings to quantify the \emph{expected} energy reduction when generating $k$ code candidates and selecting the most energy-efficient solution, given the possibility of code generation failures.

Across all applications and devices, LASSI-EE achieves $\sim$2.8× the expected energy reduction of vanilla LLM at $k=1$ (29.1\% vs. 10.5\%), 1.8× at $k=3$ (48.2\% vs. 26.3\%), and 1.6× at $k=5$ (54.6\% vs. 34.6\%). \emph{These gains emerge from both higher pass rates and deeper energy reductions produced by the LASSI-EE pipeline}.

\begin{figure}[t]
\centering
\includegraphics[width=1.0\columnwidth]{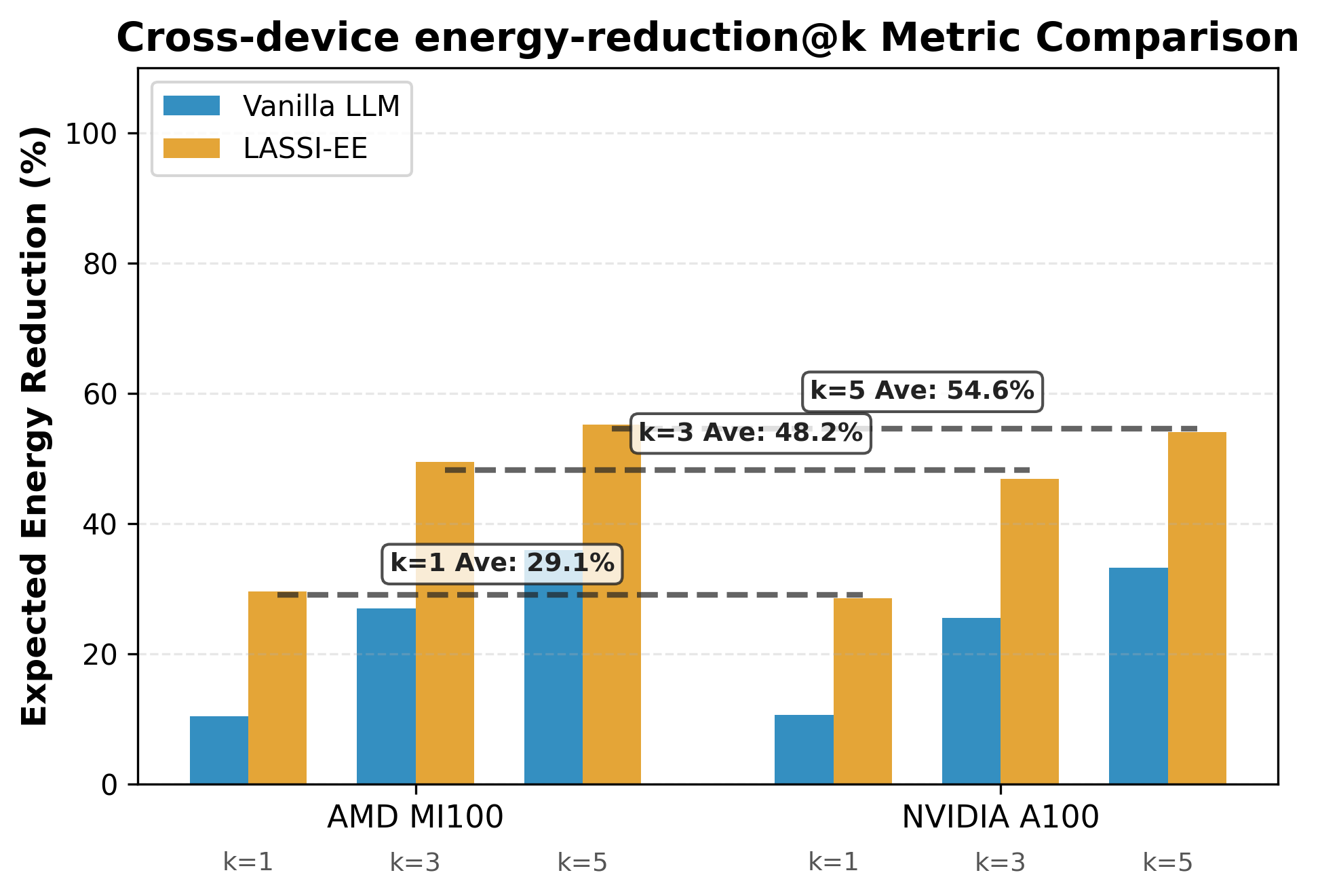}
\caption{Cross-device expected energy-reduction@$k$ results averaged across all 22 applications. LASSI-EE (orange) achieves similar gains on AMD MI100 and NVIDIA A100 over vanilla LLM (blue) for each $k\in \{1, 3, 5\}$.
}
\label{fig:e-r-at-k-cross-device}
\end{figure} 

\textsl{Cross-Device Consistency.} Small differences between average energy-reduction@$k$ values for AMD MI100 and NVIDIA A100 show consistent LASSI-EE performance across GPUs.  Figure \ref{fig:e-r-at-k-cross-device} illustrates these cross-device results, showing LASSI-EE's  improvement over vanilla LLM across both GPU architectures at $k\in \{1, 3, 5\}$, with the average savings of 29.1\% at $k = 1$, 48.2\% at $k = 3$, and 54.6\% at $k=5$, demonstrating consistent results with differences between devices of only 1 pp, 2.6 pp, and 1.1 pp, respectively.

\textsl{Diminishing Returns Beyond $k=3$.} LASSI-EE increases its average expected energy savings by +19.1 pp from $k=1$ to $k=3$, corresponding to significant +22 pp average gains over vanilla LLM at $k=3$. While improvement continues from $k=3$ to $k=5$, passing a notable threshold of $>$50\% average expected energy savings, this is a smaller jump of +6.4 pp that requires additional overhead. This result identifies $k=3$ as a favorable operating point, where LASSI-EE delivers the majority of the energy reduction observed at $k=5$ without incurring the 67\% additional code generation cost.

\textsl{Application-Specific Results.} Expected energy reductions by LASSI-EE at $k=1$ ranged from 1.0\% (\texttt{keogh} on A100) to 88.8\% (\texttt{colorwheel} on MI100) and at $k=3$ from 3.7\% (\texttt{marchingCubes} on MI100) to 99.5\% (\texttt{pathfinder} on MI100). Fourteen of 22 applications achieved $>$20\% expected energy reduction at $k=3$ on both devices. Applications with $<$10\% reductions at $k=3$ include \texttt{keogh} on A100 (5.9\%), \texttt{entropy} on A100 (6.5\%), \texttt{murmurhash3} on MI100 (9.6\%), and \texttt{marchingCubes} on MI100 (3.7\%). LASSI-EE also struggled with \texttt{jaccard} on MI100 where, despite delivering strong optimizations of up to 63.3\% energy savings at $k=5$, underperformed vanilla LLM by 11.3 pp. 

\textsl{Finding Savings Where Vanilla LLM Cannot.} While \texttt{keogh} and \texttt{entropy} on A100 show modest energy-reduction@$k$ values, these cases highlight a key strength of LASSI-EE: vanilla LLM achieved zero energy reduction for both applications at all values of $k$, despite producing valid code. LASSI-EE found real improvements (1.0\% and 3.2\% at $k=1$, respectively), with gains increasing at higher $k$. 

These results demonstrate LASSI-EE's value beyond improving pass rates: the context-building and iterative refinement pipeline uncovers optimizations that zero-shot generation misses entirely.

\section{Related Work}



Early studies of LLMs and parallel programs documented capability gaps ~\cite{chenevaluating2021, godoy_evaluation_2023, Nichols_hpc-coder_2024}. Nichols et al.~\cite{Nichols2024} observed that while LLMs (up to GPT-4) generated parallel codes with speedup over serial counterparts, they did not maximize performance. This suggested that LLMs can write parallel code but not performant parallel code that fully utilizes compute resources. LASSI-EE addresses this gap through iterative refinement guided by real-time energy measurements rather than relying solely on zero-shot LLM generation.


Recent parallel code generation efforts include 
CodeRosetta~\cite{tehranijamsaz_coderosetta_2024} for HPC language translation (CUDA, Fortran, C++) using static similarity metrics. 
Ranasinghe et al.'s~\cite{ranasinghe_llm-assisted_2025} work on code translation with FORTRAN $\rightarrow$ C++ migration and Zhu et al.'s MIRACLE~\cite{zhu_semi-supervised_2024} using back-translation for fine-tuning both evaluate on compilation and static output matching. 
Schmitz et al.'s~\cite{schmitz_parallel_2024} Parallel Pattern Language framework applies compile-time optimizations via a Domain Specific Language. 

Lou and Muller~\cite{lou_automatic_2024} optimized CUDA kernels through static resource analysis (COpPER/RaCUDA), achieving 2--4\% gains via \emph{predicted} execution costs, instead of \emph{measured} costs as in LASSI-EE. 
Finally, the  PCEBench~\cite{chen_pcebench_2025} benchmark evaluates parallel code generation across functional correctness, performance, and OpenMP/MPI support. LASSI-EE's compilation and execution pipeline is well-suited to incorporate PCEBench tasks in future scalable assessments.


A growing body of work applies LLMs to automated kernel optimization and generation for performance~\cite{zhang_accelopt_2026, novikov_alphaevolve_2025, hong_autocomp_2025, ouyang_kernelbench_2025, li_cudal1_2025, baronio_kevin_2025}, primarily on ML operator kernels. LASSI-EE targets energy efficiency on parallel scientific applications across GPU vendors.

Unlike prior studies focused on correctness, static analysis, or performance alone, LASSI-EE integrates runtime power profiling with iterative LLM-guided refinement to deliver energy-efficient parallel codes with double-digit gains. To our knowledge, this is the first study to investigate energy-aware LLM-based refactoring for parallel scientific codes. 

Previous work on energy-efficient application development and on  autotuning for performance and energy in applications and systems \cite{Wu25, Wu21, Wu16, Wu21a, Wu19, Wu17, ansel_opentuner_2014, basuroy_bliss_2021, tapus_harmony_2002, vanwerkhoven_kerneltuner_2019, nugteren_cltune_2015} often requires significant expertise in the target applications and systems, along with substantial manual tuning effort. LASSI-EE provides automatic refactoring without manual tuning effort and without the knowledge about the applications and systems for users.

\section{Conclusion}


We presented LASSI-EE, an automated LLM-based framework that generates energy-efficient parallel codes through iterative refinement guided by runtime power measurements and tailored to the target execution platform. By integrating system-aware prompting, real-time power profiling, self-correcting feedback loops, and LLM-as-a-Judge code screening, LASSI-EE achieves average energy reductions of 36\% on AMD MI100 and 34\% on NVIDIA A100 across the trials in which it produces passing energy-reducing refactorings. These results are consistent across both GPU platforms, demonstrating that LLMs can systematically generate energy-efficient parallel codes when guided by empirical execution feedback.
Notably, LASSI-EE found energy savings for applications where vanilla LLM achieved zero reduction, suggesting that context building and iterative refinement uncover optimizations that zero-shot generation misses. Our optimization analysis further shows LASSI-EE selecting different strategies across memory hierarchy, algorithmic, parallelism, and device-specific categories for different application-hardware combinations.

For practical deployment under the non-determinism of LLM inference, we introduced energy-reduction@$k$ as a metric for characterizing expected energy savings across multi-attempt code generations. Rather than requiring 30 or more runs, we identified $k=3$ as a favorable operating point that delivers the majority of the energy reduction observed at $k=5$ without incurring the 67\% additional code generation cost. We will release LASSI-EE, along with the associated datasets, as open-source on GitHub to support reproducibility and foster collaboration within the HPC and AI communities.

\section*{Acknowledgment}

...

\end{document}